\documentclass[10pt]{article}

\usepackage[letterpaper,margin=1in]{geometry}
\usepackage{mathptmx}                 
\usepackage[scaled=0.92]{helvet}      
\usepackage[T1]{fontenc}
\usepackage{microtype}

\usepackage{amsmath,amssymb}
\usepackage{booktabs}
\usepackage{array}
\usepackage{multirow}
\usepackage{makecell}
\usepackage{colortbl}
\usepackage{enumitem}
\usepackage{fontawesome5}
\usepackage{graphicx}
\usepackage{tikz}
\usepackage{tcolorbox}
\tcbuselibrary{skins,breakable}
\usetikzlibrary{arrows.meta,positioning,calc,shapes.geometric,backgrounds,fit,decorations.pathreplacing}
\usepackage[round]{natbib}
\usepackage{eso-pic}

\definecolor{arxviolet}{HTML}{6D28D9}
\definecolor{arxvioletlt}{HTML}{8B5CF6}
\definecolor{arxink}{HTML}{1E1B2E}
\definecolor{arxgreen}{HTML}{15803D}
\definecolor{arxred}{HTML}{B91C1C}
\definecolor{arxamber}{HTML}{B45309}
\definecolor{arxgray}{HTML}{6B7280}
\definecolor{arxbg}{HTML}{F5F3FF}
\definecolor{arxbg2}{HTML}{EDE9FE}
\definecolor{arxcream}{HTML}{FFF7ED}   
\definecolor{arxpeach}{HTML}{FED7AA}

\usepackage{titlesec}
\titleformat{\section}{\large\sffamily\bfseries\color{arxink}}{\thesection}{0.6em}{}
\titleformat{\subsection}{\normalsize\sffamily\bfseries\color{arxink}}{\thesubsection}{0.5em}{}
\titleformat{\subsubsection}{\small\sffamily\bfseries\color{arxink}}{\thesubsubsection}{0.5em}{}
\titlespacing*{\section}{0pt}{1.4ex}{0.8ex}
\titlespacing*{\subsection}{0pt}{1.1ex}{0.5ex}

\newtcolorbox{callout}[1][]{
  colback=arxbg, colframe=arxviolet, boxrule=0.7pt, arc=2pt,
  left=8pt,right=8pt,top=6pt,bottom=6pt, fonttitle=\bfseries\color{white},
  coltitle=white, colbacktitle=arxviolet, #1}

\newtcolorbox{algobox}[1][]{
  colback=white, colframe=arxink, boxrule=0.8pt, arc=1pt,
  left=8pt,right=8pt,top=6pt,bottom=6pt, #1}

\newtcolorbox{masthead}{enhanced, colback=arxcream, colframe=arxpeach, boxrule=1pt,
  arc=6pt, left=16pt, right=16pt, top=13pt, bottom=13pt, width=\textwidth,
  drop fuzzy shadow=arxpeach}

\newtcolorbox{promptbox}[1][Evaluation Prompt (LLM Judge)]{
  enhanced, breakable, colback=arxbg, colframe=arxviolet, boxrule=0.8pt, arc=3pt,
  left=10pt, right=10pt, top=8pt, bottom=8pt,
  coltitle=white, colbacktitle=arxviolet, fonttitle=\bfseries,
  attach boxed title to top left={xshift=10pt, yshift=-\tcboxedtitleheight/2},
  boxed title style={colback=arxviolet, arc=2pt, boxrule=0pt, left=6pt, right=6pt},
  title={\faGavel\;\,#1}, drop fuzzy shadow=arxbg2}

\newcommand{\arx}{\textsc{PaperClaw}}

\newcommand{\stage}[1]{\textbf{\color{arxviolet}#1}}
\newcommand{\ptr}{\emph{propose\,$\rightarrow$\,test\,$\rightarrow$\,reflect}}
\newcommand{\yes}{\textcolor{arxgreen}{\faCheck}}
\newcommand{\no}{\textcolor{arxred}{\footnotesize\faTimes}}
\newcommand{\tpart}{\textcolor{arxamber}{$\sim$}}

\usepackage{hyperref}
\usepackage{url}
\hypersetup{
  colorlinks=true, linkcolor=arxviolet, citecolor=arxviolet,
  urlcolor=arxviolet, filecolor=arxviolet,
  pdftitle={PaperClaw: Harnessing Agents for Autonomous Research and Human-in-the-Loop Refinement},
  pdfauthor={Weiwei Ye, Hangchen Liu, Dongyuan Li, Renhe Jiang}}

\begin{document}


\thispagestyle{plain}
\noindent
\begin{masthead}
{\centering
  \includegraphics[height=17mm]{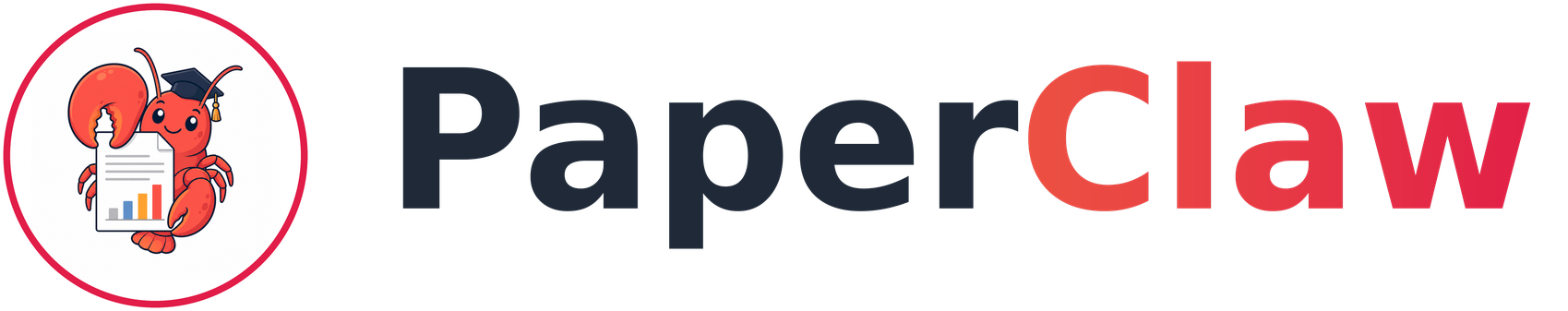}\par}
\vspace{7pt}
{\sffamily\bfseries\LARGE\color{arxink}
 Harnessing Agents for Autonomous Research\\
 and Human-in-the-Loop Refinement\par}
\vspace{8pt}
{\large Weiwei Ye \quad Hangchen Liu \quad Dongyuan Li \quad Renhe Jiang$^{*}$\par}
{\small\color{arxgray} The University of Tokyo \quad$\cdot$\quad $^{*}$Corresponding author\par}
{\small\color{arxgray}\texttt{\{wwye, liuhc3, lidy, jiangrh\}@csis.u-tokyo.ac.jp}\par}
\vspace{9pt}
\hrule height 0.4pt
\vspace{8pt}
{\small
Large language models have become capable reasoners and tool users that write and
run code and search the literature, which makes automating the research process
itself a realistic goal. We present \arx{}, a harnessed multi-agent system that
carries a project autonomously, from a field of study to a finished paper. \arx{} curates a \emph{domain} from a field's live literature,
datasets, and code; brainstorms it into an \emph{idea} with a pre-registered
main-result contract; and drives a \emph{stoppable hypothesis map} through an
iterative propose, test, reflect loop that grows only from measured verdicts and
halts once the evidence supports the idea, at which point it writes a venue-compliant
paper. A full-lifecycle memory keeps each stage in a single living record, so a long
run can be paused, inspected, and resumed without losing context. At the centre is an
\emph{in-cycle research assistant} with research tools and skills: it can drive the
whole pipeline on its own, while the same interface lets a person step in at any
stage, turning a first autonomous draft into a stronger paper through
human-in-the-loop refinement. Throughout, \arx{} keeps its output grounded and
checkable, citing only references validated against open scholarly indexes and
reporting results that genuinely ran. An evaluation with an LLM judge finds that \arx{} produces strong papers both fully autonomously and with human-in-the-loop refinement.\par}
\vspace{8pt}
\centerline{\small\faGithub~~Code: \url{https://github.com/SequenxAI/PaperClaw}}
\centerline{\small\faGlobe~~Project: \url{https://sequenxai.github.io/PaperClaw}}
\end{masthead}
\vspace{8pt}

\section{Introduction}
Large language models have grown from fluent text generators into capable reasoners
and tool users. Prompted in context, they follow detailed instructions and few-shot
examples \citep{brown2020language}, reason through a problem step by step
\citep{wei2022chain}, and call external tools, code, and search to act on the world
\citep{schick2023toolformer,yao2023react}. Each of these abilities, in-context
prompting, chain-of-thought reasoning, and tool and code use, was first studied
largely on its own. Taken together, they let a model move beyond answering a single
question toward carrying out an extended task that unfolds over many steps. This
shift, from generating an answer to executing a procedure, is what makes automating
research itself a realistic goal.

Building on these abilities, a wave of autonomous research systems now chains them
into end-to-end pipelines. The earliest take an idea through experiments to a
drafted paper \citep{lu2024aiscientist}. Others specialise: some generate and rank
research ideas \citep{si2024can,baek2024researchagent}, some execute experiments in
code or in the laboratory \citep{schmidgall2025agent,boiko2023autonomous}, and some
coordinate the work through collaborating agents \citep{gottweis2025towards}. The
most recent efforts connect several of these stages into a single self-improving
loop that turns a starting prompt into a complete draft
\citep{liu2026autoresearchclaw}. Together they show that an automated pipeline can,
in favourable cases, carry a study a long way with little human intervention.

We present \arx{}, a \emph{harnessed multi-agent system} that runs autonomous
research across the entire lifecycle, captured in its tagline, \stage{Domain
$\rightarrow$ Idea $\rightarrow$ Paper, autonomously}. It begins by curating a
domain, performing auto domain management and brainstorming ideas across one or more
fields, and then distils a chosen direction into a living idea specification. From
there it drives an iterative \ptr{} loop over a hypothesis map
(Figure~\ref{fig:method}), running and managing real experiments and growing the map
only as verdicts come in. The whole project, its domains, ideas, hypotheses, and
results, is preserved in a full-lifecycle memory, so the system can resume after
interruptions and reuse what it has already learned. Finally it writes a
venue-compliant paper, and the same capabilities are exposed identically across a
web app, a desktop app, and a command-line interface.

\arx{} provides an \emph{in-cycle research assistant}: an agent equipped with
research tools and skills and backed by the project's memory, working at every stage
of the cycle rather than only at the start. Its tools let it
survey the literature, write and run experiment code, inspect the results, and draft
prose, while its skills package common research routines so they can be reused across
hypotheses and projects. The full-lifecycle memory keeps the assistant grounded in
context: the domain it is working in, the living idea specification, the hypotheses
tried so far, and the findings each one produced. A user can step in at any point,
for example to refine the domain, propose a hypothesis, ask for an explanation of a
result, or request a draft, and the assistant folds that guidance back into the loop.

\textbf{Contributions.} \arx{} makes three contributions:
\begin{enumerate}[leftmargin=1.4em,itemsep=2pt,topsep=2pt]
\item \textbf{A clean research pipeline.} \arx{} casts autonomous research as one
clean pipeline that mirrors how a researcher works, \stage{Domain $\rightarrow$ Idea
$\rightarrow$ Hypothesis $\rightarrow$ Paper}, in a single system. It
curates a domain and brainstorms ideas from a field's live literature, datasets, and
code, pulled from open scholarly indexes \citep{priem2022openalex,lo2020s2orc}
rather than from model memory, so ideation stays current and grounded
\citep{si2024can,baek2024researchagent,wang2024scimon,yang2024large}. From there it
carries the project through to a venue-compliant paper that cites only validated
references and reports results that genuinely ran
(\S\ref{sec:model}--Appendix~\ref{sec:paper}; Table~\ref{tab:compare}).
\item \textbf{A stoppable iterative hypothesis map.} The core of the cycle is a
\ptr{} loop that grows a hypothesis map only from measured verdicts, building on
inference-time reasoning and self-correction
\citep{wei2022chain,yao2023tree,shinn2023reflexion,madaan2023selfrefine}. The loop
is \emph{stoppable}: it tracks how much of the idea's main-result contract the
evidence now supports and halts on its own once that is enough, choosing a target
venue and writing the paper rather than iterating without end (\S\ref{sec:loop}).
\item \textbf{An in-cycle research assistant.} \arx{}
provides an assistant with research tools and skills, from literature search to
running and managing experiments and drafting prose
\citep{yang2024sweagent,wang2025openhands,huang2024mlagentbench}, backed by a
full-lifecycle memory over domain, idea, hypothesis, and paper
\citep{packer2024memgpt,park2023generative,zhong2024memorybank,wang2023voyager}. A
user can step in to refine the domain, propose a hypothesis, explain a result, or
draft the paper, and the guidance is folded back into the run; throughout, output
stays grounded and checkable, with no fabricated result or citation
\citep{ji2023survey,min2023factscore}
(\S\ref{sec:loop},~Appendix~\ref{sec:impl},~Appendix~\ref{sec:memory}).
\end{enumerate}

\textbf{Positioning.} \arx{} proposes \emph{eleven} capabilities that together
span the research lifecycle: an \emph{end-to-end pipeline}, \emph{auto domain
management}, \emph{multi-domain idea brainstorm}, \emph{hypothesis-map iteration},
\emph{real-experiment execution}, \emph{experiment monitoring}, an \emph{in-cycle
research assistant}, \emph{memory evolution}, \emph{writing-style management},
\emph{multiple interfaces} (web/desktop/CLI), and an \emph{open-source}
implementation.
Table~\ref{tab:compare} maps these capabilities across representative systems,
including \textsc{AutoResearchClaw} \citep{liu2026autoresearchclaw}. Each capability
has precedents in prior work; \arx{}'s aim is
to bring them together in one open pipeline organised around a hypothesis map that
grows one verdict at a time.

\begin{table}[t]
\centering\small
\setlength{\tabcolsep}{5pt}
\renewcommand{\arraystretch}{1.2}
\resizebox{\textwidth}{!}{%
\begin{tabular}{@{}l ccccccc@{}}
\toprule
\textbf{Capability} &
\makecell{AI\\Scientist} & \makecell{AI Co-\\Scientist} & \makecell{Agent\\Lab} &
\makecell{Research-\\Agent} & \makecell{Claw\\AI Lab} & \makecell{Auto-\\Research-\\Claw} &
\textbf{\arx{}}\\
\midrule
End-to-end pipeline          & \yes & \no  & \yes & \no  & \yes & \yes & \yes\\
Auto domain management       & \no  & \no  & \no  & \no  & \no  & \no  & \yes\\
Multi-domain idea brainstorm & \no  & \no  & \no  & \no  & \no  & \no  & \yes\\
Hypothesis-map iteration     & \tpart& \tpart& \no  & \tpart& \no  & \tpart& \yes\\
Real-experiment execution    & \yes & \no  & \yes & \no  & \yes & \yes & \yes\\
Experiment monitoring        & \no  & \no  & \no  & \no  & \yes & \tpart& \yes\\
In-cycle research assistant  & \no  & \no  & \no  & \no  & \no  & \no  & \yes\\
Memory evolution             & \no  & \yes & \tpart& \no  & \tpart& \yes & \yes\\
Writing-style management     & \no  & \no  & \no  & \no  & \no  & \no  & \yes\\
Multiple interfaces (web/desktop/CLI) & \no  & \no  & \no  & \no  & \tpart& \no  & \yes\\
Open-source                  & \yes & \no  & \yes & \yes & \yes & \yes & \yes\\
\bottomrule
\end{tabular}}
\caption{The eleven proposed capabilities of \arx{} compared against
autonomous-research systems: AI Scientist \citep{lu2024aiscientist}, AI
Co-Scientist \citep{gottweis2025towards}, Agent Laboratory
\citep{schmidgall2025agent}, ResearchAgent \citep{baek2024researchagent}, Claw AI Lab
\citep{wu2026clawailab}, and \textsc{AutoResearchClaw}
\citep{liu2026autoresearchclaw}. \yes~= supported, \tpart~= partial, \no~= not
supported, based on each system's primary reported capabilities. By
\emph{in-cycle research assistant} we mean an assistant embedded at \emph{every
stage of the iterative research cycle}: a provider-agnostic scaffold that can swap
the underlying model and \emph{plug in agents} (including delegating an experiment to an external headless coding agent) rather than a one-shot, out-of-cycle
assistant.
\textsc{AutoResearchClaw} is a multi-agent debate pipeline built around a
self-healing executor; \arx{} takes a different route, organising the work around a
curated domain and a hypothesis map grown one verdict at a time, with an in-cycle
assistant and full-lifecycle memory.}
\label{tab:compare}
\end{table}

\noindent
The rest of the paper situates \arx{} against prior systems (\S\ref{sec:related}),
presents the system and its method (\S\ref{sec:paperclaw}). The experiment runners, paper compiler, full-lifecycle memory,
evolving assistant, system design, and trustworthiness safeguards are developed in
the appendices (Appendices~\ref{sec:exec}--\ref{sec:trust}).

\section{Related Work}
\label{sec:related}
\textbf{Autonomous research agents and AI scientists.} The closest systems aim
at end-to-end discovery: paper-writing pipelines \citep{lu2024aiscientist},
autonomous chemistry agents \citep{boiko2023autonomous,bran2024chemcrow},
research-assistant and co-scientist agents
\citep{baek2024researchagent,schmidgall2025agent,gottweis2025towards}, novelty and
open-domain hypothesis generators \citep{wang2024scimon,yang2024large}, and
discovery benchmarks and environments
\citep{majumder2024discoverybench,jansen2024discoveryworld,si2024can}, against the
broad backdrop of AI-for-science \citep{wang2023scientific}. Another autonomous-research
system, \textsc{AutoResearchClaw} \citep{liu2026autoresearchclaw}, is a multi-agent
self-reinforcing pipeline organised around structured multi-agent debate and a
self-healing executor with a pivot/refine loop. \arx{} takes a different approach,
organising the work around a curated domain and a stoppable hypothesis map grown
only from measured verdicts, kept in a full-lifecycle memory and closed by a
deterministic compile\,$\rightarrow$\,review loop to a venue-compliant document
(Table~\ref{tab:compare}).

\textbf{Tool-using, multi-agent and coding agents.} The experiment runners build
on tool use and program-aided reasoning
\citep{schick2023toolformer,gao2023pal,chen2023program,patil2024gorilla,qin2024toolllm,shen2023hugginggpt,nakano2021webgpt},
on embodied and grounded agents
\citep{wang2023voyager,huang2022language,ahn2022do}, on multi-agent frameworks
\citep{park2023generative,wu2023autogen,hong2024metagpt,li2023camel,qian2024chatdev,chen2024agentverse,wang2024survey,xi2025rise},
and on software-engineering agents, code models, and benchmarks
\citep{yang2024sweagent,wang2025openhands,xia2024agentless,jimenez2024swebench,huang2024mlagentbench,chen2021evaluating,li2022competition,austin2021program}.
\arx{} specialises this lineage to one goal: a defensible result record for a single
pre-registered hypothesis.

\textbf{Memory, retrieval and scholarly infrastructure.} \arx{}'s full-lifecycle
memory contrasts with conversational and episodic agent memory
\citep{packer2024memgpt,park2023generative,zhong2024memorybank} and with
retrieval-augmented generation
\citep{lewis2020retrieval,guu2020realm,karpukhin2020dense,borgeaud2022improving,asai2024selfrag,gao2024retrieval}:
where those recall dialogue or inject passages at inference time, \arx{} persists
the structured \emph{scientific} state of a project. Citations are validated against
open scholarly indexes \citep{priem2022openalex,lo2020s2orc,cohan2020specter};
science-specific models \citep{taylor2022galactica} and the literature on LLM
evaluation
\citep{hendrycks2021measuring,srivastava2023beyond,liang2023holistic,zheng2023judging}
inform the design. Finally, \arx{}'s safeguards (Appendix~\ref{sec:trust}) respond directly
to documented failure modes, hallucination and unfaithful generation
\citep{ji2023survey,huang2025survey,maynez2020faithfulness,min2023factscore}, and
to the methodological literature on pre-registration and the unreliability of
unchecked findings
\citep{nosek2018preregistration,ioannidis2005most,baker2016reproducibility,opensci2015estimating}.

\section{The PaperClaw System}
\label{sec:paperclaw}

\subsection{Method Overview}
\label{sec:overview}
\begin{figure}[t]
\centering
\includegraphics[width=\textwidth]{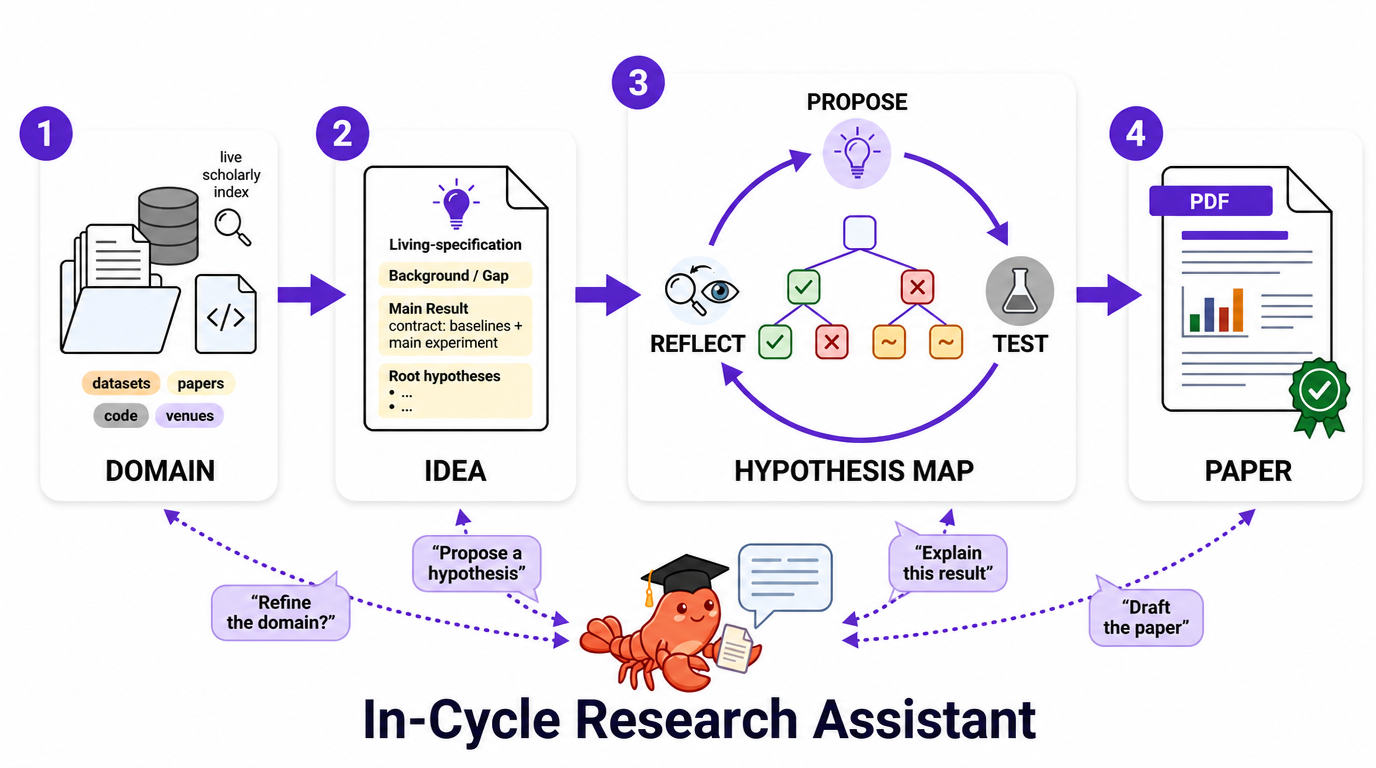}
\caption{Method overview. \arx{} turns a curated \textbf{domain} (papers,
datasets, code, venues) into an \textbf{idea} specification, then drives a
\textbf{hypothesis map} through the iterative \ptr{} loop, growing the map only
from measured verdicts (green supported, red refuted, amber inconclusive), until
the evidence is sufficient to compile a \textbf{paper}. Throughout, an in-cycle
research assistant lets a user step in at any stage, for example to refine the
domain, propose a hypothesis, explain a result, or draft the paper.}
\label{fig:method}
\end{figure}

\arx{}'s architecture is deliberately simple: it mirrors the loop a human research
group runs (Figure~\ref{fig:method}). First, a discussion that draws on in-domain
or cross-domain expertise surfaces an \emph{idea}. Second, the idea is decomposed
into \emph{hypotheses}, each verified by experiments. Third, once enough hypotheses
are positively supported, the system chooses a target venue and writes a
\emph{paper}. We develop the system in three parts, one per contribution: a clean
research pipeline (\S\ref{sec:model}), a stoppable iterative hypothesis map
(\S\ref{sec:loop}), and an in-cycle research assistant (\S\ref{sec:harness}).

\begin{figure}[t]
\centering
\begin{tikzpicture}[
  font=\small,
  lvl/.style={rounded corners=4pt, draw=arxviolet, very thick, fill=arxbg,
              minimum height=0.95cm, align=center, text width=3.0cm},
  art/.style={rounded corners=2pt, draw=arxgray, fill=white, align=left,
              text width=7.4cm, minimum height=0.95cm, font=\scriptsize, inner sep=5pt},
  down/.style={-{Latex[length=2mm]}, very thick, arxvioletlt},
  up/.style={-{Latex[length=2mm]}, very thick, arxgreen, dashed},
]
\node[lvl] (d) {\textbf{Domain}};
\node[lvl, below=0.55cm of d] (i) {\textbf{Idea}};
\node[lvl, below=0.55cm of i] (h) {\textbf{Hypothesis}};
\node[lvl, below=0.55cm of h] (p) {\textbf{Paper}};
\node[art, right=0.5cm of d] (da) {\texttt{DOMAIN.md} $\cdot$ literature, datasets, code $\cdot$};
\node[art, right=0.5cm of i] (ia) {\texttt{IDEA.md} $\cdot$ baselines, initial hypothesis    };
\node[art, right=0.5cm of h] (ha) {\texttt{.hypothesis\_map.json} $\cdot$ per-hypothesis \texttt{results.json}, \texttt{report.md}};
\node[art, right=0.5cm of p] (pa) {\texttt{paper.tex} / \texttt{paper.pdf} (versioned)};
\foreach \a/\b in {d/i,i/h,h/p}{\draw[down] ($(\a.south)+(-0.6,0)$) -- ($(\b.north)+(-0.6,0)$);}
\foreach \a/\b in {p/h,h/i,i/d}{\draw[up] ($(\a.north)+(0.6,0)$) -- ($(\b.south)+(0.6,0)$);}
\node[font=\scriptsize\itshape, arxvioletlt, anchor=east] at ($(d.south)!0.5!(i.north)+(-0.85,0)$) {reads};
\node[font=\scriptsize\itshape, arxgreen, anchor=west] at ($(h.south)!0.5!(p.north)+(0.85,0)$) {writes back};
\begin{scope}[on background layer]
\node[draw=arxink, dashed, rounded corners=5pt, fit=(d)(pa), inner sep=9pt] {};
\end{scope}
\end{tikzpicture}
\caption{Full-lifecycle memory. Each level owns a canonical, persisted record.
Stages \textcolor{arxvioletlt}{\textbf{read down}} the hierarchy (context is never
re-elicited) and \textcolor{arxgreen}{\textbf{write results back up}} (verdicts,
findings, terms). Every record is a resumable checkpoint, so an autonomous run
survives restarts and crashes.}
\label{fig:memory}
\end{figure}
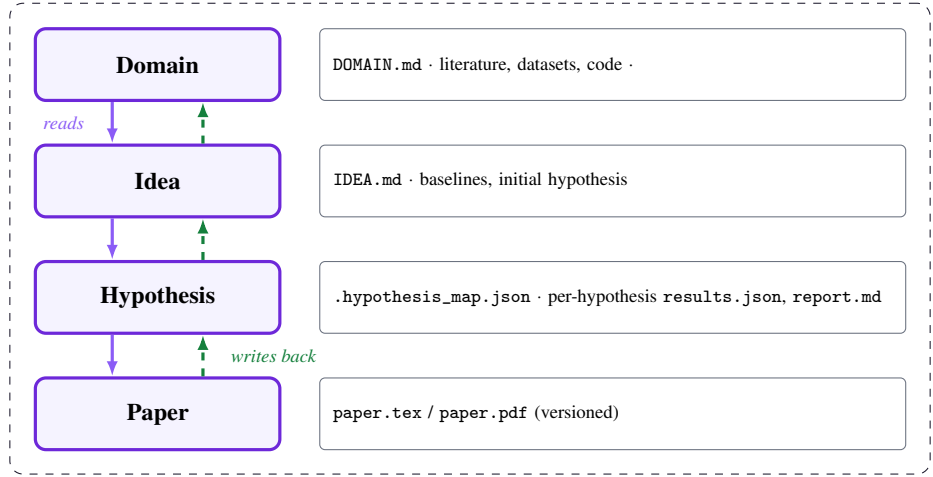

Across all four stages, \arx{} maintains a single persistent workspace with one record per level (Figure~\ref{fig:memory}): each stage reads the records above it and writes its results back, and every record is a resumable checkpoint.

Driving every stage of this loop is an \emph{in-cycle research assistant}: the agent
that does the work at each step, equipped with a library of research tools and skills
and grounded in the shared memory. Its tools span the whole cycle. To ground the
work, it searches open scholarly indexes for literature, datasets, and code, and
verifies each citation against its source. To run experiments, it writes and edits
code, launches and monitors real training or analysis jobs, reads back metrics and
logs, and can hand a heavy run to an external coding agent. To produce the paper, it
turns results into figures and tables, drafts and revises LaTeX, compiles the
document, and checks it for venue compliance. Skills package these routines so that
common tasks are reused across hypotheses and projects, and because the assistant
runs inside the loop, a user can interrupt at any stage to steer it.

\subsection{Research pipeline and memory design}
\label{sec:model}
\arx{} is \emph{idea-oriented}: every artifact descends from a concrete, testable
research direction. The pipeline moves through four stages, domain, idea, hypothesis,
and paper, mirroring how a research group works, and each stage owns a single
canonical record in a \emph{full-lifecycle memory} (Figure~\ref{fig:memory}). The
pipeline and the memory are two views of one project: the pipeline is what the system
does, the memory is what it keeps. We take each stage in turn and then describe how
the memory ties them together.

\textbf{Domain.} The pipeline starts from a curated \emph{domain}: a field's
general target, its foundational and recent papers, key datasets and benchmarks,
standard libraries, an optional canonical codebase, and its venues. The
recent-literature section is populated by a \emph{live query to an open scholarly
index} \citep{priem2022openalex,lo2020s2orc} rather than from model memory, which
keeps ideation current and avoids confabulated references. The domain is stored as \texttt{DOMAIN.md}, with its reference \texttt{codebase/} alongside, so every later stage reads the
same grounding and experiments can build on working code rather than an empty
directory \citep{jimenez2024swebench,huang2024mlagentbench}.

\textbf{Idea.} Brainstorming digests the domain into complete \emph{idea} drafts,
fully populated specifications rather than one-line sparks, which the user can refine
and promote; grounding each draft in the curated domain and live literature targets
the documented tendency of LLM ideation to be generic and to over-state novelty
\citep{si2024can,wang2024scimon}. A promoted idea owns a \emph{living specification}
that the assistant maintains throughout the project: its background and research gap,
the \emph{root hypotheses} that seed the map, and, fixed up front, a \emph{Main
Result} contract naming the baselines and the main experiment (dataset, comparison,
primary metric, and target outcome). The specification lives in \texttt{IDEA.md} and its references in a validated \texttt{ref.bib}, and fixing the Main Result before any result is seen applies
pre-registration at the level of the whole project.

\textbf{Hypothesis.} Each idea is decomposed into \emph{hypotheses} that
experiments can settle. They live in \texttt{.hypothesis\_map.json}, whose every node records a statement, a
status, and the experiment that tested it, alongside a per-hypothesis workspace
holding the plan, code, \texttt{results.json}, and a short \texttt{report.md}. This map is the memory
of what has been tried and what it showed, and it is what the iterative loop grows
one verdict at a time (\S\ref{sec:loop}).

\textbf{Paper.} Once the accumulated evidence supports the Main Result, the
pipeline compiles a \emph{paper} from the idea specification and the map's findings,
citing only the validated bibliography and reporting only results that ran. The compiled \texttt{paper.tex} and \texttt{paper.pdf} are versioned as the project's final record, so the paper
stays consistent with the evidence behind it (Appendix~\ref{sec:paper}).

\textbf{Memory management.} These files are the project's memory and together form a
single workspace the whole system reads and writes: \texttt{DOMAIN.md} holds the
grounding and \texttt{codebase/} the reference implementation, \texttt{IDEA.md} the
living specification and \texttt{ref.bib} its validated references,
\texttt{.hypothesis\_map.json} the evidence (with each hypothesis owning a
\texttt{results.json} and a \texttt{report.md}), and \texttt{paper.tex} with
\texttt{paper.pdf} the manuscript. Each stage reads the files above it, so earlier
context is never re-elicited, and writes its results back into its own file as they
arrive. Each file is a human-readable living document that the assistant rewrites
atomically rather than an append-only log, so the state a user sees is exactly the
state the engine acts on, and each doubles as a resumable checkpoint that lets a long
run be paused, inspected, or restarted after a crash without losing its place. The
full layout and reconstruction guarantees are in Appendix~\ref{sec:memory}.

\subsection{A stoppable iterative hypothesis map}
\label{sec:loop}
The scientific substance of \arx{} lives in the \emph{hypothesis map} and the
loop that drives it. The idea's root hypotheses seed the map, a hierarchy of
testable claims, rendered as a graph the user can inspect, pin, and edit. Each
node is a claim; each edge records that a child was \emph{motivated by} its
parent's measured result. The loop walks this structure, tests claims, and grows
it.

\textbf{The design principle.}
Two design choices in this section deserve their rationale stated plainly. First,
why a \emph{map} rather than a chain or a fixed outline? Because a research program
is branching and contingent: the right second question depends on the answer to
the first. Reasoning methods that lay out a chain or search a tree of
\emph{thoughts} \citep{wei2022chain,yao2023tree,besta2024graph} operate within a
single inference and never touch reality; \arx{}'s map is grown by \emph{measured
verdicts}, so its shape is a record of what the world actually said. Second, why a
\emph{loop with reflection} rather than a single planning pass? Because deciding
when enough evidence exists is itself a judgement that should be revisited as
evidence arrives, the same insight behind self-reflective and self-correcting
agents \citep{shinn2023reflexion,madaan2023selfrefine,gou2024critic}, lifted from
the level of a single answer to the level of a whole project.

\textbf{Falsifiable, pre-registered hypotheses.}
Every node in the map is a single falsifiable claim, paired with the smallest
experiment that could settle it and with explicit criteria, fixed in advance, for
when it counts as supported or refuted. Pinning those criteria before the experiment
runs is pre-registration at the level of each hypothesis, and the bar is set to be
publishable but realistic: a consistent, statistically reliable improvement on the
primary metric is enough, rather than dominance across every metric and dataset, and
the threshold may not be loosened once results are seen. The verdict is then drawn
from a fixed vocabulary, so a modest but robust gain is recorded as a genuine result
rather than dismissed. This calibration is deliberately two-sided, and it is the crux
of making an autonomous writer trustworthy: it guards against over-claiming a fragile
or imagined gain and, just as importantly, against over-refuting a real but modest
effect, mirroring a careful researcher who is skeptical of their own claims yet
willing to report a small honest result as a result.

\textbf{The status-driven loop.}
The pipeline is \emph{status-driven}: it walks the map in depth-first order and
processes any node still marked untested. Each node is handled in a dedicated
workspace through three gated steps. \stage{Plan} produces a resource estimate and
a \textsc{Feasible}/\textsc{Infeasible} decision against the detected hardware.
\stage{Experiment} runs the test and produces a structured result record, logs,
and figures. \stage{Report} issues a verdict with evidence-backed findings and
proposed follow-ups. The node's status is then set from the report's verdict, and
the loop continues. Algorithm~\ref{alg:loop} states the procedure.

\begin{algobox}[title={}]
{\small\textbf{Algorithm 1}\quad The iterative hypothesis loop}\label{alg:loop}
\hrule\vspace{4pt}
{\footnotesize\ttfamily
seed the hypothesis map from the idea's root hypotheses (roots only)\\
\textbf{for} k = 1 \textbf{to} budget:\\
\hspace*{1.5em}h $\leftarrow$ next untested node (depth-first)\quad \textbf{if} none: \textbf{stop}\\
\hspace*{1.5em}plan(h) $\rightarrow$ feasibility check\quad \textbf{if} infeasible: mark h blocked; \textbf{continue}\\
\hspace*{1.5em}experiment(h) $\rightarrow$ structured results, logs, figures\\
\hspace*{1.5em}report(h) $\rightarrow$ verdict $\in$ \{Supported, Partially, Refuted, Inconclusive\}\\
\hspace*{1.5em}status(h) $\leftarrow$ verdict\\
\hspace*{1.5em}\textbf{if} verdict is positive: expand(h) by $\le$2 motivated sub-hypotheses\\
\hspace*{1.5em}\textbf{if} reflect(map) = "enough for a paper": \textbf{stop}\\
\textbf{end for}\\
select the strongest supported subset $\rightarrow$ write $\rightarrow$ compile $\rightarrow$ review $\rightarrow$ refine
}
\end{algobox}

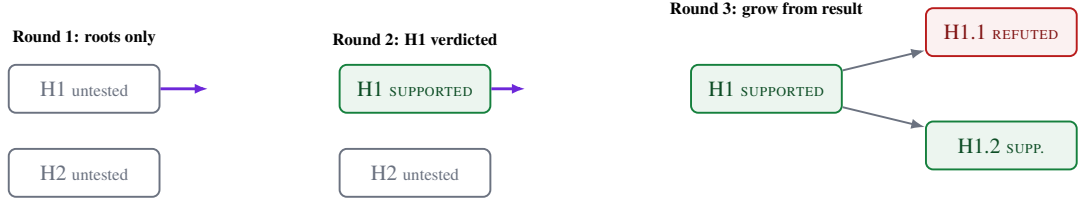
\begin{figure}[t]
\centering
\resizebox{0.86\textwidth}{!}{%
\begin{tikzpicture}[
  font=\small,
  hyp/.style={rounded corners=3pt, draw, thick, minimum width=2.2cm,
              minimum height=0.7cm, align=center},
  unt/.style={hyp, draw=arxgray, fill=white, text=arxgray},
  sup/.style={hyp, draw=arxgreen, fill=arxgreen!8, text=arxgreen!60!black},
  ref/.style={hyp, draw=arxred, fill=arxred!6, text=arxred!70!black},
  arr/.style={-{Latex[length=2mm]}, thick, arxgray},
]
\node[unt] (a1) {H1 \scriptsize untested};
\node[unt, below=0.5cm of a1] (a2) {H2 \scriptsize untested};
\node[font=\scriptsize\bfseries, above=0.15cm of a1] {Round 1: roots only};

\node[sup, right=2.6cm of a1] (b1) {H1 \scriptsize\textsc{supported}};
\node[unt, below=0.5cm of b1] (b2) {H2 \scriptsize untested};
\node[font=\scriptsize\bfseries, above=0.15cm of b1] {Round 2: H1 verdicted};

\node[sup, right=2.9cm of b1] (c1) {H1 \scriptsize\textsc{supported}};
\node[ref, above right=0.1cm and 1.2cm of c1] (c11) {H1.1 \scriptsize\textsc{refuted}};
\node[sup, below right=0.1cm and 1.2cm of c1] (c12) {H1.2 \scriptsize\textsc{supp.}};
\node[font=\scriptsize\bfseries, above=0.55cm of c1] {Round 3: grow from result};
\draw[arr] (c1) -- (c11);
\draw[arr] (c1) -- (c12);

\draw[-{Latex[length=2mm]}, very thick, arxviolet] (a1.east) -- ++(0.7,0);
\draw[-{Latex[length=2mm]}, very thick, arxviolet] (b1.east) -- ++(0.5,0);
\end{tikzpicture}%
}
\caption{The hypothesis map grows \emph{one verdicted level at a time}. Children
may be added only after the parent has an experimental verdict, so H1.1 and H1.2
are motivated by H1's measured result, never invented up front. A refuted branch
(H1.1) simply stops; a supported one (H1.2) can be expanded further within the
budget.}
\label{fig:map}
\end{figure}

\textbf{Growing the map one verdict at a time.}
A single rule defines the character of \arx{}'s research: \emph{generation creates
root nodes only}, and the map grows one verdicted level at a time. The engine
refuses to add children under a parent whose status is still untested or
blocked, a sub-hypothesis must be \emph{motivated by its parent's measured
result}, never invented in advance (Figure~\ref{fig:map}). After a positive
verdict, at most two child hypotheses are proposed and themselves tested within the
budget; after a refutation, that branch simply stops, and attention moves
elsewhere. Nodes receive hierarchical identifiers (H1, then H1.1, H1.2, and so on)
and each owns its own workspace, so the provenance of every claim (what motivated it, what tested it, and what it concluded) is preserved.

A reflection step after each round asks a deliberately blunt question: is the
evidence accumulated so far \emph{enough for a paper}? When the answer is yes, the
loop stops early; when it is no, the loop continues until the budget is exhausted.
This makes the procedure adaptive to how quickly evidence accumulates: an idea
whose first two hypotheses land cleanly can finish in a handful of rounds, while a
harder idea explores more of its map before concluding.

\subsection{An in-cycle research assistant}
\label{sec:harness}

\begin{figure}[!h]
\centering
\includegraphics[width=\textwidth]{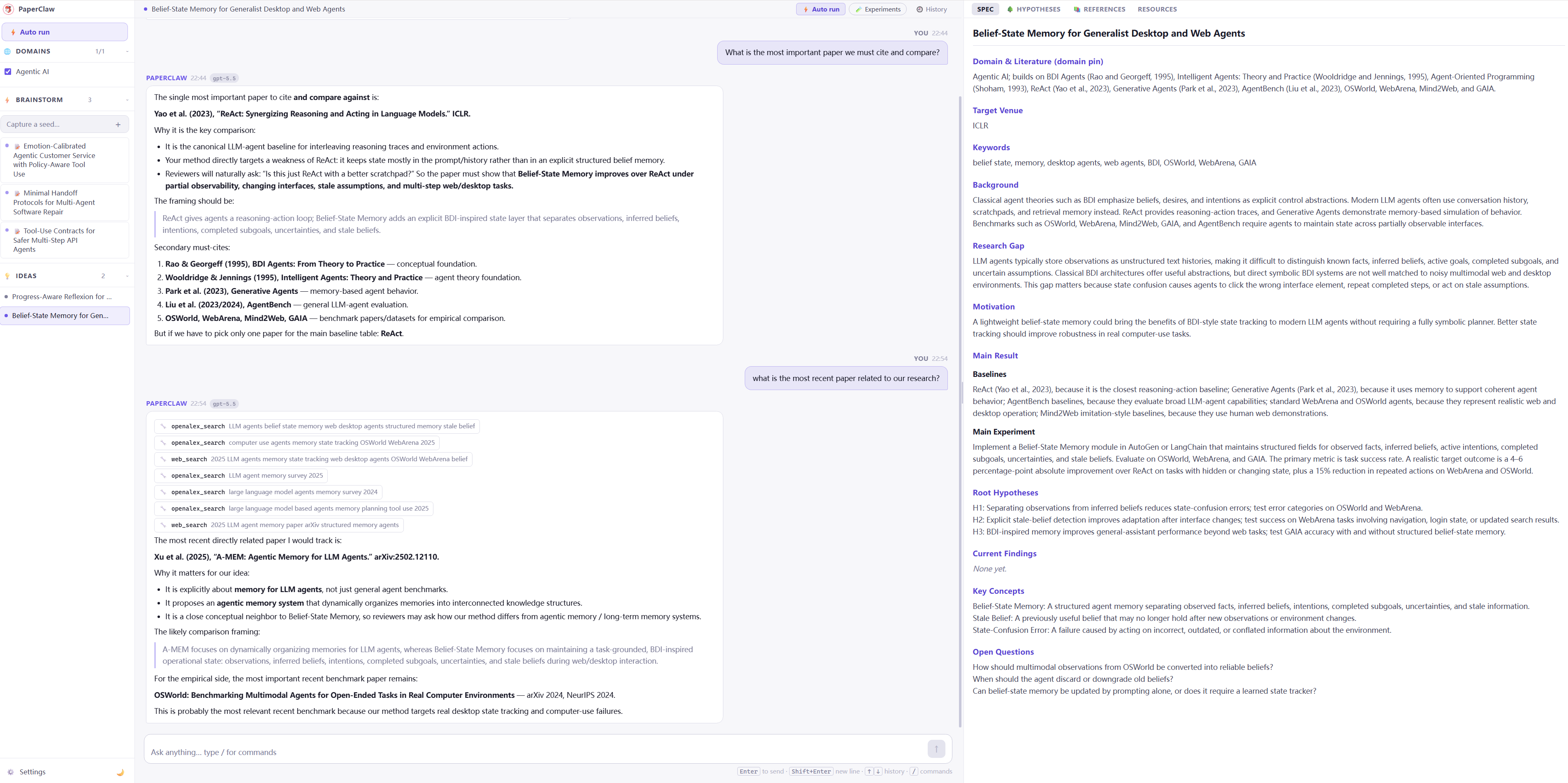}
\caption{The in-cycle research assistant in human-in-the-loop use. The chat (left)
lets a user collaborate with the assistant on the live project, here proposing and
discussing hypotheses with one-click tools and skills, while the compiled paper
updates alongside (right). The same interface also runs unattended in autonomous
mode.}
\label{fig:assistant}
\end{figure}

Throughout the loop, the in-cycle research assistant introduced in
\S\ref{sec:overview} continuously prompts and improves the code, the experiments,
and the paper, while \arx{} itself owns the control flow (the hypothesis map, the
experiment monitoring, and the memory). The agent doing the reasoning at each step
is a component that can be substituted or upgraded, so a stronger agent can drive the
same loop without changing the surrounding system
(Appendix~\ref{sec:exec},~Appendix~\ref{sec:impl}).

\textbf{Tools and skills.} The assistant acts through two layers
(Table~\ref{tab:toolskill}). \emph{Tools} are low-level primitives it calls directly
to read and edit workspace files, search the web and OpenAlex, fetch pages, read PDFs
and images, record validated citations, and edit the hypothesis map. \emph{Skills}
are higher-level named operations, exposed as slash commands, that each orchestrate
tools and the model to carry out one step of the pipeline. A user can invoke any
skill explicitly, and the system calls the same skills on its own during an
autonomous run.

\begin{table}[t]
\centering\small
\setlength{\tabcolsep}{6pt}\renewcommand{\arraystretch}{1.25}
\begin{tabular}{@{}p{0.40\textwidth} p{0.52\textwidth}@{}}
\toprule
\textbf{Tool} & \textbf{What it does}\\
\midrule
\texttt{read\_file}, \texttt{write\_file}, \texttt{apply\_patch}, \texttt{list\_files} & read, create, patch, and list workspace files\\
\texttt{read\_pdf}, \texttt{read\_image} & read a compiled PDF's rendered text; view a figure\\
\texttt{web\_search}, \texttt{fetch\_url} & search the public web and fetch a page's text\\
\texttt{openalex\_search}, \texttt{cite} & find real papers and append a validated citation to \texttt{ref.bib}\\
\texttt{hypothesis\_add/update/remove} & add, edit, or delete nodes of the hypothesis map\\
\midrule
\textbf{Skill} & \textbf{What it does}\\
\midrule
\texttt{/create\_domain}, \texttt{/setup\_codebase} & build a domain and attach a reference codebase\\
\texttt{/pin\_idea}, \texttt{/idea\_generation} & turn a brainstorm or conversation into an idea (\texttt{IDEA.md})\\
\texttt{/generate\_hypothesis\_map} & generate the hypothesis map (\texttt{.hypothesis\_map.json})\\
\texttt{/generate\_plan}, \texttt{/generate\_report} & write a hypothesis's test plan, then its report and follow-ups\\
\texttt{/validate\_references} & check every \texttt{ref.bib} entry against Crossref and OpenAlex\\
\texttt{/setup\_venue}, \texttt{/write\_paper} & fetch the venue template, then write the paper in a chosen style\\
\bottomrule
\end{tabular}
\caption{Key tools and skills of the in-cycle research assistant. \emph{Tools} are
primitives the agent calls directly; \emph{skills} are named research operations
(slash commands) that orchestrate tools and the model. The same skills run on demand
from a user or autonomously by the system.}
\label{tab:toolskill}
\end{table}

\textbf{A study of the assistant.} Figure~\ref{fig:assistant} shows the assistant at
work on a live project. The chat on the left is where a user and the assistant
collaborate: the assistant proposes and discusses hypotheses, each backed by a
one-click skill, while the compiled paper updates alongside on the right. The two
questions in the transcript illustrate where its answers come from. The first it
answers directly from the project's memory, reading the domain and idea records
(\texttt{DOMAIN.md} and \texttt{IDEA.md}) without re-eliciting context. The second
needs information not yet in memory, so it reaches for its tools, a web search and a
scholarly-search query to OpenAlex, to pull in current literature before replying. The
same screen thus exposes the whole loop in one place: the memory it reads, the tools
and skills it calls, and the evolving paper it is writing.

\textbf{Human-in-the-loop refinement.} Because the assistant runs inside the loop, the
same interface that can run unattended also lets a person steer it at every gate, and
this human guidance is what lets a paper improve over time rather than in a single
shot. A user can accept or revise a proposed answer, redirect the next hypothesis, ask
for an explanation of a result, or request a redraft, and each change is folded
straight back into the run, so the manuscript is developed continuously rather than
produced once. The ablation in \S\ref{sec:experiments} measures the effect on paper
quality, and we see this continuous, human-in-the-loop development, with the system
carrying the routine work and a researcher supplying judgement at the decision points,
as a central direction for future work.

\section{Experiments}
\label{sec:experiments}
We evaluate the quality of the papers each system produces, scoring them with an LLM
judge (Claude) against a fixed rubric. This section reports our evaluation; the judging rubric and prompt are detailed in
Appendix~\ref{app:protocol}.

\subsection{Setup}
\textbf{Papers and systems.} We score the example papers from their GitHub pages. For
The AI Scientist \citep{lu2024aiscientist}, \textsc{AutoResearchClaw}
\citep{liu2026autoresearchclaw}, and Agent Laboratory \citep{schmidgall2025agent} we
take their GitHub example papers; for a \emph{naive single-LLM writer} baseline, which
ships no examples, we generate papers by prompting a single model to write the code
and paper directly, with no agentic loop; and for \arx{} we use the example papers
from our own GitHub page.

\textbf{Judging.} Each paper is rated on a 1 to 10 scale along four qualitative
dimensions (Novelty, Soundness, Experimental rigor, and Clarity), plus a holistic
Overall, by an LLM judge (Claude) that sees the paper with its system identity hidden,
following the rubric anchors and prompt in Appendix~\ref{app:protocol} and the
LLM-as-judge protocol \citep{zheng2023judging}.

\subsection{Results}
Table~\ref{tab:mainresults} reports an LLM-judge (Claude) assessment of each
system's available example papers. \arx{}, run in human-in-the-loop mode, attains
the highest overall score, pairing clear writing with real benchmarks, multiple
baselines, ablations, and bootstrap confidence intervals. The other systems each
show their own strengths: the AI Scientist's papers are focused and self-contained,
\textsc{AutoResearchClaw}'s are clear and well-scoped across diverse fields, and the
naive single-LLM writer produces fluent, readable drafts. The main axis of
difference is how much measured experimental evidence each paper carries.

\begin{table}[t]
\centering\small
\setlength{\tabcolsep}{6pt}\renewcommand{\arraystretch}{1.25}
\begin{tabular}{@{}l cccc c@{}}
\toprule
\textbf{System} & Novelty & Soundness & \makecell{Exp.\\rigor} & Clarity & \textbf{Overall}\\
\midrule
Naive single-LLM writer & 2.5 & 5.0 & 4.0 & 7.3 & 4.0\\
The AI Scientist \citep{lu2024aiscientist} & 3.9 & 4.9 & 4.8 & 6.8 & 4.8\\
Agent Laboratory \citep{schmidgall2025agent} & 3.0 & 4.0 & 4.0 & 7.0 & 4.0\\
\textsc{AutoResearchClaw} \citep{liu2026autoresearchclaw} & 4.3 & 4.5 & 3.0 & 7.3 & 4.3\\
\textbf{\arx{} (ours)} & \textbf{6.2} & \textbf{7.0} & \textbf{7.5} & \textbf{8.0} & \textbf{7.0}\\
\bottomrule
\end{tabular}
\caption{Main results: end-to-end paper quality by system on a 1 to 10 scale
(higher is better). Scores are an LLM-judge (Claude) assessment of each system's
available example papers. Novelty, Soundness, Experimental
rigor, and Clarity follow the rubric anchors (Table~\ref{tab:anchors}); Overall is
a holistic recommendation score. The \arx{} entry is generated in human-in-the-loop
mode (\S\ref{sec:experiments}, ablation).}
\label{tab:mainresults}
\end{table}

\subsection{Ablation: human-in-the-loop refinement}
\arx{}'s answer-first interaction lets a human steer the system at every gate.
Starting from the fully autonomous draft, a human provides continuous, targeted
feedback (accepting or revising the assistant's proposed answers and requesting
specific improvements), and the system re-runs only the affected stages.
Table~\ref{tab:hitl} scores a fully autonomous \arx{} run and a human-in-the-loop
run, on different topics, with the same judge. On these papers the two tie at
Overall 7.0: when an autonomous run carries its experiments through to completion,
it already produces a strong, rigorous paper, so human guidance is not always
needed to reach that quality. The benefit of human-in-the-loop refinement shows up
instead when an autonomous run stalls; a separate autonomous draft that reached only
a feasibility check, with no measured results, scored far lower (Overall 3.0). Human
steering therefore mainly lifts the weakest autonomous runs rather than the
strongest.

\begin{table}[t]
\centering\small
\setlength{\tabcolsep}{6pt}\renewcommand{\arraystretch}{1.25}
\begin{tabular}{@{}l ccccc@{}}
\toprule
\textbf{Configuration} & Novelty & Soundness & \makecell{Exp.\\rigor} & Clarity & \textbf{Overall}\\
\midrule
\arx{} (autonomous) & 6.0 & \textbf{7.0} & 7.5 & 7.1 & 6.5\\
\textbf{\arx{} (ours)} & \textbf{6.2} & \textbf{7.0} & \textbf{7.5} & \textbf{8.0} & \textbf{7.0}\\
\bottomrule
\end{tabular}
\caption{Human-in-the-loop ablation. A completed autonomous \arx{} paper and a
human-in-the-loop \arx{} paper, scored by the same judge (different topics, not a
matched pair). They tie at Overall 7.0; human steering matters most when an
autonomous run stalls, where a separate autonomous draft with no measured results
scored 3.0.}
\label{tab:hitl}
\end{table}

\section{Conclusion}
\arx{} harnesses autonomous agents across the entire research lifecycle. By
managing domains and brainstorming ideas across them; by driving an iterative
hypothesis loop whose reasoning is supplied by an in-cycle, swappable research
assistant; by running and managing real experiments; by preserving and reusing the
whole project in a full-lifecycle memory; and by compiling a venue-compliant
document that cites only validated work, the system turns ``write me a paper'' into
a structured, resumable, and reusable pipeline, delivered identically across web,
desktop, and command line, and unifying eleven capabilities that prior systems
provide only in part. The remaining gap to a fully trustworthy AI scientist is,
fittingly, an \emph{empirical} one: replacing simulation with execution everywhere,
isolating that execution, and evaluating the scientific quality, not just the
compliance, of what comes out. \arx{} is built so that closing that gap changes the
plugged-in agents, not the surrounding system.

\bibliography{references}
\clearpage
\appendix
\section{Experimental Protocol}
\label{app:protocol}
This appendix details how the papers compared in \S\ref{sec:experiments} are scored.

\textbf{Judge.} Each paper is scored by a single LLM judge (Claude) at temperature
$0$ for determinism.

\textbf{Anonymisation.} Before judging, every paper is re-rendered into one uniform
template and stripped of author names, acknowledgements, repository URLs, and any
system self-identification (including the string \arx{} and competitor names), so the
judge cannot infer the source from style or metadata.

\textbf{Rubric.} The judge rates four dimensions on a $1$ to $10$ scale: Novelty
(originality of the question and approach relative to prior work), Soundness
(correctness of method and claims, absence of overclaiming), Experimental rigor
(appropriate baselines, ablations, statistics, and reproducibility), and Clarity
(organisation, writing, and figures), plus a holistic Overall recommendation. The
shared score anchors are in Table~\ref{tab:anchors}; a system's score is the mean of
its papers' scores, and Overall is a separate holistic judgement rather than the mean
of the four dimensions. The exact judge prompt is shown below; the placeholder in
double braces is filled per paper.

\begin{table}[!h]
\centering\small
\renewcommand{\arraystretch}{1.2}
\begin{tabular}{@{}cl@{}}
\toprule
\textbf{Score} & \textbf{Anchor (applies to each dimension)}\\
\midrule
9--10 & Excellent: publishable as is on this dimension; no substantive weakness.\\
7--8  & Good: solid, with minor weaknesses a revision would fix.\\
5--6  & Borderline: notable gaps that a reviewer would require addressing.\\
3--4  & Weak: serious problems that undermine the contribution.\\
1--2  & Poor: absent, incorrect, or unusable on this dimension.\\
\bottomrule
\end{tabular}
\caption{Shared $1$ to $10$ scoring anchors used by the judge on each rubric
dimension.}
\label{tab:anchors}
\end{table}

\begin{promptbox}
\small
You are an expert reviewer for a top venue. You are given a \textbf{paper} to
assess. Judge the paper's \emph{content} only; do not reward length or polish, and
do not penalise short papers that are correct. Be strict and well calibrated.

\medskip
\noindent\texttt{PAPER:\ \{\{paper\_text\}\}}

\medskip
\noindent\textbf{Answer these questions, grounding every judgement in the paper:}
\begin{enumerate}\itemsep2pt
  \item \textbf{Novelty.} Is the question or method original relative to the prior
        work it cites and to what you know?
  \item \textbf{Soundness.} Are the methods correct and the claims supported by the
        evidence, with no overclaiming?
  \item \textbf{Experimental rigor.} Are there appropriate baselines, ablations,
        and statistics, and enough detail to reproduce the results?
  \item \textbf{Clarity.} Is the paper well organised and readable, and are the
        figures and tables informative?
  \item \textbf{Overall.} Taking everything together, would you recommend
        acceptance?
\end{enumerate}

\noindent\textbf{Scoring.} Give \texttt{idea\_fidelity} as a boolean, then score
\textbf{Novelty}, \textbf{Soundness}, \textbf{Experimental rigor}, \textbf{Clarity},
and \textbf{Overall} each from $1$ to $10$ using the anchors in
Table~\ref{tab:anchors}. Justify each score in one sentence tied to the paper.
Return \emph{only} JSON in exactly this form:

\medskip
{\ttfamily\footnotesize\setlength{\parindent}{0pt}%
\{\\
\hspace*{1.5em}"idea\_fidelity": true | false,\\
\hspace*{1.5em}"novelty":\ \ \ \ \{"score": 1-10, "reason": "..."\},\\
\hspace*{1.5em}"soundness":\ \ \{"score": 1-10, "reason": "..."\},\\
\hspace*{1.5em}"rigor":\ \ \ \ \ \ \{"score": 1-10, "reason": "..."\},\\
\hspace*{1.5em}"clarity":\ \ \ \ \{"score": 1-10, "reason": "..."\},\\
\hspace*{1.5em}"overall":\ \ \ \ \{"score": 1-10, "reason": "..."\}\\
\}\par}
\end{promptbox}

\section{Grounded Experiment Execution}
\label{sec:exec}
A hypothesis is only as credible as the experiment that tests it. \arx{} therefore
treats the experiment phase as a \emph{pluggable} component, selected by a
run-configuration setting, so the same loop scales from a zero-setup demonstration
to a genuine compute run behind one interface (Table~\ref{tab:runners}). This
separation is deliberate: it lets the scientific logic of the loop (propose, test, reflect, grow) remain fixed while the \emph{fidelity} of the test is dialled up or
down to match the resources and trust available.

\begin{table}[t]
\centering\small
\renewcommand{\arraystretch}{1.3}
\begin{tabular}{@{}>{\raggedright\arraybackslash}p{1.9cm} >{\raggedright\arraybackslash}p{7.1cm} >{\raggedright\arraybackslash}p{3.0cm}@{}}
\toprule
\textbf{Mode} & \textbf{What it does} & \textbf{Use case} \\
\midrule
\textsc{simulated} & The model narrates plausible, mixed results under strict
anti-fabrication prompts. Works with zero setup and no compute. &
Demonstration, ideation. \\
\textsc{executed} & A real multi-file coding agent builds a clean module layout
through write, patch, and shell actions, runs the \emph{real} datasets, and
iterates until it produces a structured result record. &
Local CPU/GPU, real measurements. \\
\textsc{remote} & The same generate--run--fix loop pushes code to a configured
remote machine, runs it there, and pulls the results and figures back. &
Cluster / remote compute. \\
\textsc{delegated} & Hands the whole experiment to an external headless
coding-agent command, streaming its output live and reading the result record on
exit. & Reusing a best-in-class external agent. \\
\bottomrule
\end{tabular}
\caption{The pluggable experiment runners. A single selector routes every call
site through one of them, so a hypothesis can be tested by a narrated simulation or
by code that genuinely runs, without changing the surrounding loop.}
\label{tab:runners}
\end{table}

\subsection{The coding agent}
In executed mode the agent drives a thinking-augmented loop in which each step may
emit several actions at once. It builds a real codebase (separate modules for data handling, the model, training, and evaluation) and edits them surgically with
context-matched patches that are located by matching the surrounding lines rather
than by trusting line numbers, which models routinely get wrong. It runs commands
and reads their output, then revises, iterating until it produces a structured,
machine-readable result record and any figures. Because native tool-calling is not
universally streamable across providers, the action protocol is carried in plain
text, which lets the agent's thinking, code, and command output stream live for
\emph{both} major provider families, an important property for a long-running job
a human may want to watch. The design is squarely in the lineage of tool-using and
software-engineering agents \citep{schick2023toolformer,yang2024sweagent,wang2025openhands,xia2024agentless},
but specialised to one narrow goal: produce a defensible result record for a single
pre-registered hypothesis, not solve an open-ended task.

\subsection{Detached, monitored execution}
Real experiments can run for hours, so \arx{} executes them as \emph{detached,
monitored jobs}. Each experiment runs in its own process, independent of the user
interface, and emits an append-only event stream plus a status record. The
application itself is a thin \emph{monitor}: it tails the event stream and checks
whether the process is alive, so a run survives a restart of the interface and can
be re-attached afterward, and a job whose process has vanished is reconciled to an
\emph{interrupted} state rather than reported as still running. There is
deliberately \emph{no timeout}, a research run is not a web request, and killing
it at an arbitrary deadline would discard exactly the long experiments that matter
most. A single monitor view lists every job across all ideas and jumps to the live
output on demand.

\subsection{Hardware-aware feasibility gating}
Before committing compute, \arx{} detects what is available. A deterministic probe,
run identically on the local machine and on any configured remote, classifies the
processor, accelerators, memory, and disk and records a hardware snapshot. The plan
step then gates each hypothesis with a \textsc{Feasible}/\textsc{Infeasible}
decision against that snapshot: a claim whose estimated resource needs exceed what
is available is marked \emph{blocked} and its experiment is skipped, rather than
launched and left to fail hours later. A separate, fast readiness check, requiring
no model calls, verifies that the broader environment is in order (a writable
workspace, a configured model, a working typesetting toolchain, and image
generation) and reports each item as healthy, a warning, or a failure. Feasibility
gating is a small idea with a large effect on autonomy: it lets the system decline
to attempt the impossible, which is itself a form of honesty.

\section{From Findings to a Compiled, Venue-Compliant Paper}
\label{sec:paper}
When reflection judges the evidence sufficient, \arx{} writes a paper, and,
unlike a system that emits text and stops, it produces a \emph{compiled,
venue-compliant} document. The distinction matters because most of what makes a
manuscript acceptable or not at a venue (length limits, the mandated style, the resolution of every citation and cross-reference, the presence of required structural elements) is invisible until the document is actually typeset.

\subsection{Generation and the compile\,$\rightarrow$\,review\,$\rightarrow$\,refine loop}
The paper stage selects the strongest supported subset of hypotheses, generates
both conceptual and data figures, and writes the manuscript directly in the
venue's typesetting format. When an official venue template is available, the
manuscript is built on that skeleton, reusing the template's preamble
verbatim, so the locally compiled document matches the venue exactly rather than
approximating it. Prose is governed by \emph{writing-style management}: a library
of reusable style guides (kept globally and per-domain, and selectable per paper)
separates \emph{how} the manuscript reads (tone, structure, phrasing) from the
venue's formatting, so the same results can be written up in a chosen voice and a
group's preferred style carried from one paper to the next. Compilation uses a standard, full typesetting toolchain when one
is present and a self-contained engine otherwise; missing packages are resolved and
installed automatically and the build retried, so even a minimal installation
behaves like a complete one. Crucially, the agent then \emph{reads its own
compilation log back} and iterates a fix loop until the document builds cleanly,
treating a failed build as a problem to debug rather than an error to surface.

\subsection{Deterministic compliance review}
Generation is paired with a \emph{deterministic} reviewer, with no model in the
loop, so it cannot be talked out of an error or persuaded that a violation is
acceptable. The reviewer lints the source for disallowed packages, commands, and
style options; compiles the document; checks the page limit declared by the venue;
scans the build log for margin overflows, undefined citations and references, and
missing figures; and reads the finished document back to confirm that required
structural elements (such as an abstract and a reference list) are present. The
paper stage loops review\,$\rightarrow$\,fix until the report is clean, so
length and style compliance are \emph{guaranteed by construction} rather than hoped
for. This compile\,$\rightarrow$\,verify\,$\rightarrow$\,refine pattern is the
document-level analogue of the experiment loop: generate, check against an
external ground truth, and repair.

\subsection{Verified citations}
Fabricated citations are a signature failure of LLM-written papers
\citep{ji2023survey,min2023factscore}, so \arx{} never lets the model invent a
reference. Each idea owns a bibliography built \emph{deterministically} from open
scholarly indexes, with conventional, human-readable citation keys. A validator
then checks every entry independently, by identifier against one index and by
title against another, and labels it \textsc{verified}, \textsc{mismatch},
\textsc{not found}, or \textsc{unknown}, catching any fabricated or mis-attributed
reference before it can reach the manuscript. Only validated entries are supplied
to the writer, and a dedicated tool lets the agent add a real paper (resolved to a
canonical identifier) as it works. The bibliography of \emph{this} paper was
assembled and checked by exactly this procedure, which is why every reference in
it resolves to a real, indexed work.

\section{Full-Lifecycle Memory: Domain \texorpdfstring{$\rightarrow$}{->} Idea \texorpdfstring{$\rightarrow$}{->} Hypothesis \texorpdfstring{$\rightarrow$}{->} Paper}
\label{sec:memory}
A research project is not a single context window; it is a body of accumulating
state that outlives any one conversation. \arx{} therefore maintains a single
\emph{hierarchical memory} that spans the entire lifecycle, with one persistent
store per level (Figure~\ref{fig:memory}). This is not the conversational or
episodic memory that lets an agent remember a dialogue
\citep{packer2024memgpt,park2023generative,zhong2024memorybank}, nor the retrieval
that injects external passages at inference time
\citep{lewis2020retrieval,guu2020realm,karpukhin2020dense,borgeaud2022improving,asai2024selfrag,gao2024retrieval}.
It is a structured \emph{research} memory: the durable scientific state of a
project, organised exactly as the work itself is organised.

\subsection{Four levels, one store}
Each level owns a canonical record that is the single source of truth for
everything beneath it. A \textbf{domain} owns its literature, datasets, code, and
optional reference implementation. An \textbf{idea} owns its living specification
and its validated bibliography. A \textbf{hypothesis} owns its node in the map and
a dedicated workspace holding its plan, code, results, logs, and report. The
\textbf{paper} owns the compiled manuscript and its versions. Nothing of
consequence lives only in a transcript; everything that matters is written down at
the level where it belongs.

\subsection{Reading down, writing up}
Memory flows in both directions, and that bidirectionality is what makes the
hierarchy a memory rather than a mere archive. \emph{Downward}, every stage reads
the levels above it: the paper is built from verified hypothesis reports, which are
built from the idea's Main Result contract, which is anchored to the domain's
literature. Because each level reads from the one above, context is never
re-elicited from the user and never silently re-invented by the model.
\emph{Upward}, results are written back: an experiment verdict updates its node's
status in the map; a synthesised finding is appended to the idea's current
findings; a newly coined term is added to the glossary. The consequence is that the
specification the user sees is always exactly the state the engine acts on, a
living contract, not a stale summary.

\subsection{Memory makes autonomy durable}
Because every phase persists its own checkpoint, the pipeline is \emph{resumable by
construction}: re-invoking it skips any phase whose checkpoint already exists, a
round is considered finished once its reflection is written, stopping and
continuing resumes at the first incomplete phase, and an explicit restart clears
checkpoints to begin again. Combined with detached experiment jobs
(Appendix~\ref{sec:exec}), even a hard crash in the middle of a multi-hour run loses
nothing, the project is reconstructed from its persisted state alone. Durability
is not a convenience feature here; it is a precondition for autonomy, because a
process that must be supervised continuously is not autonomous at all.

\section{An Evolving Research Assistant}
\label{sec:evolve}
A one-shot system answers a request and forgets it. \arx{} is built instead to
\emph{accumulate}, so that each project leaves the assistant better equipped for
the next. Three mechanisms make this concrete, and together they turn the
full-lifecycle memory of Appendix~\ref{sec:memory} from a within-project record into a
genuinely \emph{evolving} assistant.

\textbf{Reusable knowledge assets.} Several of the artifacts the system creates
are not idea-specific; they are durable assets that persist beyond the project that
produced them and are reused by later ones. A \emph{domain} (its curated literature, datasets, libraries, and reference implementation) is built once and
then anchors any number of ideas, growing more valuable as it is corrected and
extended. \emph{Prose-style guides} that capture how a group likes to write are
saved globally and per-domain and applied to future manuscripts. A domain's
\emph{reference implementation} lets every later experiment build on working code
rather than a blank file. And a \emph{validated bibliography} accumulates verified
references that subsequent papers can draw on. Each of these is a unit of reusable
research capital, in the spirit of agents that acquire and retrieve reusable skills
\citep{wang2023voyager,packer2024memgpt}, but at the level of a field rather than a
single task.

\textbf{Improvement through human feedback.} The answer-first interaction of
\S\ref{sec:model} is not only a convenience; it is the channel through which the
assistant is corrected and steered. Every question the assistant poses carries its
own proposed answer, so a human can accept, adjust, or override it with minimal
effort, and those choices are written back into the persistent specification and
the reusable assets. Over time this feedback shapes the domains, the styles, and
the hypotheses the system favours, an inexpensive but continual form of alignment
to a particular researcher's taste, complementary to the verbal self-correction
that operates within a single run \citep{shinn2023reflexion,madaan2023selfrefine}.

\textbf{Evolution within and across projects.} Inside a project, the hypothesis
map already \emph{evolves}: it grows only where evidence licenses it, so its final
shape is a learned artifact rather than a fixed plan (\S\ref{sec:loop}). Across
projects, the reusable assets and accumulated feedback mean the assistant starts
each new idea from a richer base than the last. This is a deliberately different
stance from session-scoped evolution, such as the hypothesis tournaments of an AI
co-scientist \citep{gottweis2025towards}, which sharpen ideas within a single run. The closest related notion is the \emph{cross-run
evolution} of \textsc{AutoResearchClaw} \citep{liu2026autoresearchclaw},
which turns past mistakes into future safeguards; \arx{} pursues the same goal of a
non-amnesic assistant but through reusable, inspectable knowledge assets rather than
learned safeguards. We are explicit about the current boundary: \arx{}
\emph{reuses and is steered}. Fully closed-loop learning, in which outcomes automatically update
the policies that propose hypotheses and design experiments, is a direction the
architecture is designed to support rather than a capability we claim today.

\section{System Design}
\label{sec:impl}
\arx{} is delivered as a single, open-source application with a shared service
core, three interchangeable front ends (a web app, a desktop app, and a command-line interface) and a persistent workspace. The design goal throughout is that the
science is implemented once and exposed everywhere.

Architecturally, \arx{} is best understood as an \emph{in-cycle research
assistant}: a provider-agnostic scaffold into which the underlying intelligence is
\emph{plugged at every turn of the research cycle}, rather than a one-shot,
out-of-cycle assistant. At
each stage of the loop (plan, experiment, reflect, write) the model client is
swappable across providers (\S\ref{sec:impl-stream}); the experiment phase is a
pluggable runner that can even \emph{delegate a whole experiment to an external
headless coding agent} (Appendix~\ref{sec:exec}); and the chat agent itself is
interchangeable. \arx{} thus supplies the scientific control flow (the hypothesis map, the verdict loop, the verification and memory) while the assistant doing the
reasoning at each step is a component that can be substituted or upgraded. This is
what the \emph{in-cycle research assistant} row of Table~\ref{tab:compare} refers
to.

\subsection{One capability, many surfaces}
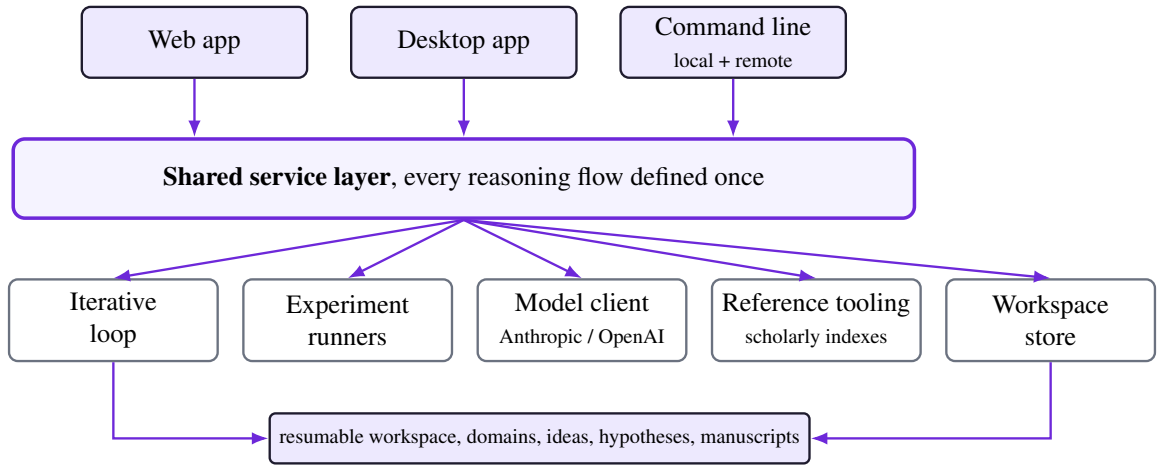
\begin{figure}[t]
\centering
\resizebox{0.92\textwidth}{!}{%
\begin{tikzpicture}[
  font=\small,
  surface/.style={rounded corners=3pt, draw=arxink, thick, fill=arxbg2,
                  minimum width=2.7cm, minimum height=0.85cm, align=center},
  core/.style={rounded corners=4pt, draw=arxviolet, very thick, fill=arxbg,
               minimum width=10.8cm, minimum height=0.95cm, align=center},
  mod/.style={rounded corners=3pt, draw=arxgray, thick, fill=white,
              minimum width=2.5cm, minimum height=0.98cm, align=center},
  darr/.style={-{Latex[length=2mm]}, thick, arxviolet},
]
\node[surface] (web) {Web app};
\node[surface, right=0.5cm of web] (desk) {Desktop app};
\node[surface, right=0.5cm of desk] (cli) {Command line\\\scriptsize local + remote};
\node[core, below=0.7cm of desk] (svc) {\textbf{Shared service layer}, every reasoning flow defined once};
\draw[darr] (web.south) -- (svc.north -| web.south);
\draw[darr] (desk.south) -- (svc.north);
\draw[darr] (cli.south) -- (svc.north -| cli.south);
\node[mod, below=0.7cm of svc, xshift=-4.2cm] (loop) {Iterative\\loop};
\node[mod, right=0.28cm of loop] (exec) {Experiment\\runners};
\node[mod, right=0.28cm of exec] (llm) {Model client\\\scriptsize Anthropic / OpenAI};
\node[mod, right=0.28cm of llm] (refs) {Reference tooling\\\scriptsize scholarly indexes};
\node[mod, right=0.28cm of refs] (store) {Workspace\\store};
\foreach \m in {loop,exec,llm,refs,store}{\draw[darr] (svc.south) -- (\m.north);}
\node[draw=arxink, thick, rounded corners=2pt, fill=arxbg2, below=0.6cm of exec, xshift=2.3cm,
      minimum width=6.4cm, minimum height=0.6cm, align=center] (disk)
      {\scriptsize resumable workspace, domains, ideas, hypotheses, manuscripts};
\draw[darr] (store.south) |- (disk.east);
\draw[darr] (loop.south) |- (disk.west);
\end{tikzpicture}%
}
\caption{Architecture. Three surfaces (web, desktop, command line) call \emph{one}
shared service layer, which orchestrates the iterative loop, the pluggable
experiment runners, a provider-agnostic model client, deterministic reference
tooling, and a persistent, resumable workspace. The simultaneity rule keeps all
three surfaces feature-identical.}
\label{fig:arch}
\end{figure}

A central rule, which we call the \emph{simultaneity rule}, requires that every
feature land on \emph{all} surfaces in the same change: the service, the graphical
front ends, and the command line. The mechanism that makes this tractable is a
shared service layer (Figure~\ref{fig:arch}): every flow that involves the model is
implemented \emph{once} in that layer and invoked identically by the graphical
clients and by the command line, which can also drive a running instance remotely.
A strict contract keeps the data types on the wire identical on both sides of the
boundary. The practical consequence is that the terminal and the graphical
interface are never out of step: the same autonomous run can be launched on one,
monitored on another, and resumed on a third. For a tool intended to run long,
unattended jobs, this is not a cosmetic nicety, it is what lets a user start a run
from a laptop and check on it from a server.

\subsection{A provider-agnostic, streaming core}
\label{sec:impl-stream}
All model calls go through one client that can speak to either a first-party model
provider or any compatible third-party endpoint, defaulting to a current frontier
model. The same code path \emph{streams the model's intermediate reasoning} as well
as its answer, so a long thinking phase is visible as it happens rather than
appearing as a frozen pause, an important affordance when a single step can take
minutes. A tool-use loop (reading and editing files, searching the literature and
the web, recording citations, editing the hypothesis map, and compiling and
reviewing the manuscript) runs uniformly across providers, with tool descriptions
translated to each provider's format on the fly. Every model-dependent feature
degrades gracefully when no provider is configured, falling back to a helpful
message rather than an error, so the application is usable (for browsing, planning, and inspection) even with no credentials at all.

\subsection{Resumability and observability}
Two cross-cutting properties recur throughout the system and are worth naming on
their own. \emph{Resumability} (Appendix~\ref{sec:memory}) means every long operation is a
sequence of checkpointed phases, so nothing is ever redone unnecessarily and
nothing is ever lost. \emph{Observability} means every long operation streams its state (thinking, generated code, command output, phase transitions) to whichever
surface is watching, and persists that stream so it can be replayed later. Together
they make a multi-hour autonomous run behave less like a black box and more like a
process a human can supervise, interrupt, inspect, and trust.

\section{Trustworthiness: Anti-Fabrication and Calibration}
\label{sec:trust}
Autonomy without discipline is a fabrication machine: a sufficiently fluent model,
asked to ``write a paper,'' will happily produce one whose every sentence is
well-formed and whose every claim is unsupported. \arx{} therefore treats
trustworthiness as a first-class design constraint, enforced at exactly the points
where a language model is most tempted to cut a corner. Table~\ref{tab:safeguards}
pairs each failure mode of autonomous paper generation with the mechanism that
prevents it; each mechanism has appeared in context above, and here we collect them
as a single contract.

\begin{table}[t]
\centering\small
\renewcommand{\arraystretch}{1.3}
\begin{tabular}{@{}>{\raggedright\arraybackslash}p{3.6cm} >{\raggedright\arraybackslash}p{8.6cm}@{}}
\toprule
\textbf{Failure mode} & \textbf{Safeguard in \arx{}} \\
\midrule
Unfalsifiable claims & Every hypothesis is a single claim with explicit
\emph{Supported if / Refuted if} criteria; superlatives and conjunctions are
rejected. \\
Moving the goalposts & Acceptance criteria are \emph{pre-registered} before the
experiment and may not be loosened after results are seen. \\
Result inflation / over-claiming & A modest, robust gain is \textsc{Supported},
not ``state of the art''; verdicts are calibrated, and \textsc{Partially
Supported} is recorded as a positive-but-qualified result. \\
Over-refuting real effects & A bounded, realistic threshold counts a reliable
small improvement as support rather than discarding it. \\
Inventing the answer up front & The map grows \emph{one verdict at a time};
children must be motivated by a measured parent result. \\
Fabricated experiments & Real runners produce a structured result record from code
that executes; even simulation is governed by explicit anti-fabrication prompts. \\
Fabricated citations & The bibliography is built and \emph{validated} against open
scholarly indexes; only verified entries reach the paper. \\
Field drift & The pipeline refuses to run an idea with no anchoring domain, keeping
the work tied to a real literature. \\
\bottomrule
\end{tabular}
\caption{\arx{} encodes scientific discipline as control flow and prompt contracts.
Each row is a way an autonomous writer can produce a plausible-but-false paper,
paired with the mechanism that prevents it.}
\label{tab:safeguards}
\end{table}

The unifying theme is that discipline is enforced \emph{structurally}, not merely
requested in a prompt. A prompt that says ``do not fabricate'' is advice the model
may or may not follow; a control flow that refuses to add a sub-hypothesis until
its parent has a verdict, or a deterministic reviewer that will not pass an
unresolved citation, is a constraint the model cannot talk its way around. Where a
behaviour genuinely depends on the model's judgement, most importantly, the
calibration of verdicts, \arx{} pins that judgement to a fixed, two-sided rubric
(\S\ref{sec:loop}) so that it neither inflates fragile gains nor dismisses real
ones. The result is not a guarantee of correct science, which no system can
provide, but a sharp reduction in the specific, well-documented ways that fluent
generation goes wrong.

\section{The Persistent Workspace and Reproducibility}
\label{app:artifacts}
The full-lifecycle memory of Appendix~\ref{sec:memory} is realised as a single persistent
workspace with one nested scope per level of the hierarchy. A \emph{domain} scope
holds the domain specification and, optionally, a reference implementation. Each
\emph{idea} scope holds its living specification, its validated bibliography, the
hypothesis map, the per-round checkpoints written by the loop, a nested scope for
each \emph{hypothesis} (containing that hypothesis's plan, code, structured
results, logs, figures, and report), and the compiled manuscript and its versions.

Reproducibility follows from two properties of this layout. First, every phase is
\emph{idempotent up to its checkpoint}: re-running the pipeline skips any phase
whose record already exists, so a run can be stopped, inspected, and continued at
will. Second, the workspace is \emph{self-contained}: a run can be reproduced,
audited, or resumed from its persisted state alone, with no reliance on
in-memory context. Together these mean that the path from a domain to a finished
paper is not an ephemeral conversation but a durable, inspectable record.

\section{The Structured-Intent Interaction Protocol}
\label{app:blocks}
The assistant communicates structured intent through tagged blocks embedded in its
otherwise natural-language replies, parsed tolerantly so that minor formatting
variation does not break them. A single streamed reply can therefore both speak to
the user in prose and act on the workspace: it may rewrite the idea specification,
create a new idea or domain, revise a draft, or pose an interactive question (a
small menu of options whose first entry is the assistant's own recommended answer,
in keeping with the answer-first rule of \S\ref{sec:model}). Each reply also carries
lightweight flags indicating what changed, whether a specification was updated, an
idea or domain created, the hypothesis map edited, or the manuscript
regenerated, so the watching interface can refresh exactly the affected view. The
design lets one model turn advance the conversation and the project state together,
which is what makes the interaction feel like collaborating with a co-author rather
than issuing commands to a tool.

\section{Extended Related Work: Foundations and Reasoning}
\label{app:related}
Beyond the autonomous-research systems compared in \S\ref{sec:related}, \arx{}
rests on broader foundations in language modelling and in inference-time
reasoning, summarised here.

\textbf{Foundations: language models and scaling.} \arx{} builds on the
transformer and the pretrain--finetune paradigm
\citep{vaswani2017attention,devlin2019bert,raffel2020exploring}, on in-context
learning and scale
\citep{brown2020language,kaplan2020scaling,hoffmann2022training,wei2022emergent},
on instruction tuning and alignment
\citep{ouyang2022training,bai2022constitutional}, and on the frontier models
\citep{chowdhery2023palm,touvron2023llama,openai2023gpt4} that supply the reasoning
substrate behind every stage.

\textbf{Reasoning, prompting and self-correction.} The \ptr{} loop draws on a
lineage of inference-time reasoning (chain-of-thought and its zero-shot and
self-consistent variants
\citep{wei2022chain,kojima2022large,wang2023selfconsistency,zhou2023least},
deliberate and compositional search
\citep{yao2023tree,besta2024graph,zhou2024selfdiscover}, and acting while reasoning
\citep{yao2023react}) and on self-correction
\citep{shinn2023reflexion,madaan2023selfrefine,gou2024critic,chen2024teaching}.
Where these operate on a single answer, \arx{}'s reflection operates on a whole
\emph{research program}, deciding whether the accumulated experimental evidence
warrants a paper.

\end{document}